\documentclass[letterpaper, 10 pt, conference]{root}  

\IEEEoverridecommandlockouts                              

\usepackage{cite}

\title{\LARGE \bf
Trust in Queer Human-Robot Interaction
}

\author{Raj Korpan$^{1}$
\thanks{$^{1}$Raj Korpan was with Iona University. He is currently with the Department of Computer Science, Hunter College, New York, NY 10065, USA.
        {\tt\small rk1665@hunter.cuny.edu}}%
}

\begin{document}

\maketitle
\thispagestyle{empty}
\pagestyle{empty}

\begin{abstract}

Human-robot interaction (HRI) systems need to build trust with people of diverse identities. This position paper argues that queer (LGBTQIA+) people must be included in the design and evaluation of HRI systems to ensure their trust in and acceptance of robots. Queer people have faced discrimination and harm from artificial intelligence and robotic systems. Despite calls for increased diversity and inclusion, HRI has not systemically addressed queer issues. This paper suggests three approaches to address trust in queer HRI: diversifying human-subject pools, centering queer people in HRI studies, and contextualizing measures of trust.

\end{abstract}

\section{INTRODUCTION}

Acceptance of robots in the wild requires systems that build trust with people with diverse cultural and racial backgrounds, ages, sexes, and abilities. This position paper argues that queer people are an important and distinct population that must be explicitly included in the design of human-robot interaction (HRI) systems to address their particular barriers to trusting and accepting robots. Furthermore, measures of trust may not accurately capture the factors affecting trust for such marginalized minority populations even if they have been validated for the general population. Novel methods developed to assess trust should address the general lack of trust in institutions and technology among queer communities.

\section{BACKGROUND}
Recent calls for increased diversity, equity, and inclusion (DEI) in artificial intelligence (AI) and robotics have described how discrimination and bias exist at multiple levels: in the implementations of systems, the research community, and the targeted users \cite{fosch2022diversity}. Sex and gender diversity are at most risk in these systems, particularly in robotics \cite{Fosch2023accounting}. To address these issues, workshops have promoted DEI in HRI \cite{Tanevska2023inclusive}. Others called for the inclusion of feminist principles \cite{Winkle2023feminist} and intersectional gender inclusivity \cite{hipolito2023enactive} in HRI practices.

There have also been specific calls to address queer issues in AI \cite{Liliequist2023beyond} and robotics \cite{Poulsen2020queering}. Queer in AI is an organization that has demonstrated the use of participatory design to both bring visibility to the harms that AI systems have on queer people and also to empower queer inclusion in the AI research community \cite{OrganizersOfQueerin2023QueerAI}. Recent work showed that advances in algorithmic fairness have not addressed queer concerns \cite{tomasev2021fairness} and found that popular AI systems do not center the experiences of transgender and non-binary individuals \cite{ovalle2023m}.

Not only are queer issues ignored in the creation of HRI systems, their impacts on queer people are also not examined. One study found that no HRI studies centered sex and gender issues across 415 papers from 2002 to 2023, and only two addressed sexuality but in the context of sex robots \cite{Seaborndiversity}. Similarly, queer people are not included as subjects in HRI studies: of 55,032 participants from 787 studies, only two trans people were reported and sexuality was only reported in one study \cite{Seaborn2023not}. Furthermore, (presumably cis-gender) men are over-represented as research participants \cite{winkle202315}. Consequently, the impact of HRI systems on queer people are often completely unmeasured or, even worse, they may cause further discrimination and harm.

One recent work addressed the use of robots for elder care with queer people \cite{poulsen2020cybersecurity,Poulsen2020value}. In that work, queer participants expressed a desire for robotic systems that prioritize privacy and their physical and digital safety, and also understand, appreciate and respect the unique differences of the queer community. That research also found that in the care-giving context, trust is especially important between the caregiver and care recipient and that similar factors affect trust with a human caregiver as with a robot caregiver \cite{poulsen2022investigation}.

A robot's gender presentation can also impact trust in HRI. One study showed that participants attributed more cognitive and affective trust to a robot with a female body shape \cite{bernotat2021fe}. Notably, however, that study did not ask participants about their gender identity or sexuality and so it may not generalize to the queer community. Moreover, the body shapes presented further stereotypes and exclude the diversity of queer bodies.

Measurement of trust is a major challenge in human-robot interaction because it cannot be directly observed \cite{kok2020trust}. Most commonly, self-report scales are used as a measurement tool for trust, such as the Muir trust questionnaire \cite{muir1990operators}, the Trust in Automation Scale \cite{jian2000foundations}, the Human-Computer Trust Instrument \cite{madsen2000measuring}, the Merritt scale \cite{merritt2011affective}, the Human-Robot Interaction Trust Scale \cite{yagoda2012you}, the Human-Robot Trust Scale \cite{Schaefer2013perception}, and the Multi-Dimensional Measure of Trust (MDMT) \cite{malle2021multidimensional}. The MDMT, for example, measures trust along five dimensions: reliability, competency, ethicality, transparency and benevolence \cite{ullman2019measuring}. These tools have not all been assessed for reliability or validated \cite{khavas2021review,Ueno2022trust}, especially on the same set of participants to determine the level of agreement across measures \cite{chita2021can}. Although others have attempted to use physiological measures to infer trust (e.g., \cite{jenkins2010measuring,akash2018classification}), it is unclear how reliable these approaches are as a measure of trust. Furthermore, none of these have been evaluated for their effectiveness and applicability to the queer community.

\section{TRUST IN QUEER HRI}
Despite calls for increased DEI in HRI, the literature shows that there has not been a sustained effort to address how systems are designed and evaluated with respect to their impact on queer people. The result is a limited understanding of how queer people interact with robots, what factors guide their acceptance and trust in these interactions, and how measures of trust could adapt to a queer audience. I offer three suggestions for how to address these issues: diversifying human-subject pools, centering queer people in HRI studies, and contextualizing measures of trust.

\subsection{DIVERSIFY HUMAN SUBJECT POOLS}
A major gap in the human-robot trust literature is the lack of human subjects with queer identities \cite{Seaborn2023not}. Visibility is the first step towards inclusion. As a starting point, I believe researchers should ask participants about their gender identity and sexuality. Instead of a lack of clarity about human subjects in HRI studies, inclusive practices like participatory design allow for better evaluation of trust with people \cite{OrganizersOfQueerin2023QueerAI}. The challenge with the collection of participants' queer characteristics is the potential for privacy violations or the misuse of this information for discriminatory or harmful purposes \cite{williams2023voice}. Despite these concerns, it is possible to build trust with vulnerable populations through intentionally designed studies that prioritize their issues \cite{lee2022configuring}.

Understanding the demographics of subjects is not enough to achieve improved DEI, a concerted effort must also be made to recruit queer individuals into HRI studies \cite{winkle202315}. Instead of confronting the challenge of recruiting a representative sample, many HRI studies use convenience samples of university students \cite{baxter2016characterising}, but the difficulty of recruitment should not be an excuse to exclude vulnerable and marginalized peoples. Without a diverse subject pool, we cannot begin to examine the impact on trust in queer HRI.

\subsection{CENTER QUEER PEOPLE}
Queer people trust technology designed for them more than systems created for the majority non-queer population \cite{beare2021queer,haimson2020designing,haimson2023transgender}. Rarely have HRI systems focused on queer needs: only Poulsen et al. pursued such work \cite{poulsen2020cybersecurity,Poulsen2020value,poulsen2022investigation}. Queer individuals face systematic discrimination and challenges in their lives distinct from non-queer people (e.g., in healthcare \cite{ayhan2020systematic} and the workplace \cite{mara2021strategies}). Robots could have a unique opportunity to support this marginalized community to address their issues, such as mental health, bullying, loneliness, lack of access to traditional systems of support \cite{mongelli2019minority}. To achieve these goals, however, HRI researchers must build systems designed for and evaluated with queer people, as proposed by Queer in AI \cite{OrganizersOfQueerin2023QueerAI}. Furthermore, the factors that affect trust will likely be different because of the discrimination queer people face (e.g., organizations face distinct challenges when engaging with queer communities \cite{ciszek2020we}). Because AI and robotics systems have already been shown to be harmful and discriminatory toward queer people \cite{Poulsen2020queering}, I believe that it is important for HRI to center queer people and their needs while not furthering harmful stereotypes.

\subsection{CONTEXTUALIZE MEASURES OF TRUST}
Although measures of trust are treated as a standardized way to evaluate HRI systems across many settings and tasks, regardless of the target population \cite{kok2020trust}, the way queer people experience trust is different from non-queer people (e.g., \cite{ciszek2020we}). While the standard approach provides consistency and reliability, it may ignore the contextualized, nuanced, and intersectional experience of queer individuals \cite{ruberg2020data}. Instead, inclusive measures of trust need to address the aspects of trust specifically relevant to the queer community, such as differences in gender identities and orientations and use of preferred pronouns and inclusive language \cite{meyer2022rainbow}. Simultaneously, these tools need to respect queer individuals' emotional and psychological well-being and privacy concerns.

The challenge with this approach is how to do this in a rigorous, verifiable, and reusable way because most HRI studies use a single trust measurement tool instead of separate demographic-specific measures. One approach could combine quantitative measures with qualitative data to provide clarity to the trust measure. Another approach is to use an iterative user-centered design process that involves queer people to produce a tool that is culturally competent \cite{fine2019critical}. A third approach could be a customizable questionnaire that includes context-based questions that address different factors that affect trust, such as the robot's appearance, performance, and behavior, the task, and the participant's identity. Although the MDMT is modular in that it provides five dimensions that can be selected based on the scenario \cite{ullman2019measuring}, it does not adapt to other contextual factors, such as a participant's queer identity. A customizable approach could combine some common trust questions with context-specific questions. Alternatively, it could weigh the importance of trust factors differently. For example, queer people face unique concerns around privacy \cite{williams2023voice} and so a trust measure could weigh the importance of that higher than other factors.

\section{CONCLUSIONS}
As robots are deployed in the real world, they need to interact with all kinds of humans. The challenge for these robots is to solve difficult tasks in unstructured dynamic environments while they are accepted and trusted by their human partners. This position paper focuses on trust in \textit{queer human}-robot interaction. Despite calls for more inclusive approaches, queer people have not been broadly considered by HRI researchers in the design or evaluation of systems. Measures of human-robot trust also fail to consider the unique factors that could affect trust with queer individuals.

I offer three suggestions toward a more queer HRI. First, a person's queer identity is likely a confounding factor in the measure of trust, so studies should both ask about queer identity and recruit more queer participants. Second, HRI systems should be developed for the needs of the queer community. Third, a person's queer identity likely affects their perception of trust, so trust measures should consider that context. These three suggestions would expand who these systems are for, understand what affect they have on queer communities, and examine how trust could be reliably measured for them. Queer people also deserve the potential benefits of robots but to achieve that we must first understand how to build trust in queer human-robot interaction.






\section*{ACKNOWLEDGMENT}

The author thanks the reviewers whose valuable comments and suggestions helped improve this paper.


\bibliographystyle{root}
\bibliography{root}

\end{document}